\documentclass[]{jdmdh}
\usepackage{array}
\usepackage{pgfplots}
\pgfplotsset{compat=1.16}
\usepackage{lscape}
\usepackage{tikz}
\usetikzlibrary{arrows.meta}
\tikzset{>={Latex[width=3mm,length=3mm]}}
\usetikzlibrary{shapes,positioning}

\usepackage{longtable}
\usepackage{microtype}

\title{Corpus and Models for Lemmatisation and POS-tagging\\ of Old French}

\author[1]{Jean-Baptiste Camps}
\author[1]{Thibault Clérice}
\author[1]{Frédéric Duval}
\author[1]{Lucence Ing}
\author[1]{Naomi Kanaoka}
\author[1]{Ariane Pinche}

\affil[1]{Centre Jean-Mabillon, École nationale des chartes, Université Paris, Sciences \& Lettres}

\corrauthor{Jean-Baptiste Camps}{Jean-Baptiste.Camps@chartes.psl.eu}

\begin{document}

\maketitle

\abstract{%
Old French is a typical example of an under-resourced historic languages, that furtherly displays an important amount of linguistic variation. In this paper, we present the current results of a long going project (2015-…) and 
describe how we broached the difficult question of providing lemmatisation and POS models for Old French with the help of neural taggers and the progressive constitution of dedicated corpora. %
}

\section{Introduction}
\label{sec:introduction}
\strut

Today, linguistically annotated corpora\footnote{%
The process of lemmatisation associates a canonical form with each term in a text, regardless of its inflection.%
} are at the centre of crucial research issues, as they constitute long-term and reusable data, ensure the reproductibility of a study, and allow the interrogation and exploitation of large-scale textual sets \citep{mellet_les_2002}. Since the 1960s, the annotation of corpora has been automated by linguists, particularly in the field of NLP (Natural Language Processing), for the quantitative exploitation of linguistic phenomena. The identification of the occurrences of a word is thus facilitated by annotation, which allows a more efficient harvesting. These corpora also constitute essential data for dialectometry or stylometry, in particular for author attribution studies \citep{mellet_lemmatisation_2002} 
or for the automatic classification of literary genre \citep{feldman_part--speech_2009}. Finally, lemmatisation can be integrated into a semi-automatic processing pipeline starting from the digitisation of a manuscript and leading to the production of a critical edition by comparing the texts of different witnesses \citep{camps_collating_2019}. The range of possibilities makes this data extremely valuable.

Lemmatisation as an academic task was first developed in the study of flectional ancient languages, such as Latin or Greek, because this computerisation of the task already corresponded to the ancient use of concordance tables or indexes \citep{riche_concordance_1984}. Roberto Busa and his team working on the lemmatisation of the \textit{Summa Theologica} of Thomas Aquinas starting in the 1950s \citep{busa_annals_1980} were pioneers in this area. Ten years later, the LASLA\footnote{%
    Laboratoire d’Analyse Statistique des Langues Anciennes, University of Liège.
} set up a system of lemmatisation of its corpora by collecting all information manually and recording this information in computer files. In the 1990's, the emergence of lemmatisers such as \textit{TreeTagger} \citep{schmid_probabilistic_1994} made it possible to automatically annotate corpora. These tools were at the time based on rules and a dictionary to predict the linguistic annotation of a word (token) in a given language. 

In the 1980s, linguistic annotation was extended to Old French with the desire to systematically study the language of a document and its dialectal particularities with the
constitution of two manually annotated corpora with a set of 225 numeric tags encoding part of speech and other morphological categorie, the first one of charters and the second one of literary texts, \textit{Amsterdam Corpus},
by Anthonij Dees (Vrije Universiteit Amsterdam) and his collaborators, including Piet van Reenen, in order to produce the two atlas of the linguistic forms of  Old French \citep{dees_atlas_1980,dees_atlas_1987}. 
While the Charter corpus has been lost, in the 2000s, the literary corpus of the second atlas was taken over and fully lemmatised and annotated (addition of POS), 
with the help of the lemmatiser \textit{TreeTagger} for Old French and using lemmas proposed in the \textit{Tobler-Lommatzsch dictionary}
\citep{tobler_altfranzosisches_2002} 
by Pierre Kunstmann and Achim Stein to form the \textit{Nouveau corpus d'Amsterdam} (NCA, \cite{kunstmann_nouveau_2007}). But, vernacular languages such as Old French bring other challenges related to spelling variations, making it difficult to use a fixed dictionary for lemmas. Indeed, the same word can have a large number of graphic realisations. For example, the term \textit{cheval} can have more than thirty different forms (see Table 1, \cite{camps_collating_2019}), which makes it pretty much inefficient to use a lemmatisation tool based on a pre-established dictionary. 

\begin{table}[htbp] \small \centering
\begin{tabular}{lr}
\textbf{Form}    & \textbf{Freq.}\\ \hline \hline
cheval  & 785 \\
cheual  & 375 \\
chevaus & 248 \\
ceval   & 98  \\ 
chevax  & 92  \\ 
chevals & 84  \\ 
ceual   & 66  \\ 
cheuaus & 65  \\ 
chival  & 34 
\end{tabular}
\begin{tabular}{lr}
\textbf{Form}    & \textbf{Freq.}\\ \hline \hline
chevaux & 30     \\  
chivaus & 27     \\ 
cheuax  & 23     \\  
chiual  & 23     \\ 
cevaus  & 19     \\
chevas  & 19     \\
cheuals & 14     \\ 
cevals  & 12     \\ 
chiuaus & 11        
\end{tabular}
\begin{tabular}{lr}
\textbf{Form}    & \textbf{Freq.}\\ \hline \hline
ceuax   & 10      \\ 
cevax   & 10      \\ 
ceuaus  & 9       \\ 
chiuau  & 9       \\      
cheuaux & 8       \\ 
kevaus  & 6       \\ 
chevau  & 5       \\ 
cevaux  & 3       \\ 
chivals & 3       
\end{tabular}
\begin{tabular}{lr}
\textbf{Form}    & \textbf{Freq.}\\ \hline \hline
cheuas  & 2       \\ 
keval   & 2       \\ 
chaval  & 1       \\
chavaux & 1       \\
cheua   & 1       \\ 
cheualx & 1       \\ 
cheuau  & 1       \\
chevalx & 1       \\ 
chiuals & 1
\end{tabular}
\caption{spellings for ``cheval'' in the \textit{Nouveau corpus d'Amsterdam}}
\label{tab:cheval}
\end{table}

Traditional lemmatisers such as \textit{Treetagger} work with rule sets, a lexicon that only recognises lemmas that already exist in its training corpus, and a decision-tree based algorithm. Due to their fixed lexicon which mainly takes into account only flectional variations and the token environment, the results are not as optimal as those of classical Latin or Modern French due to the unstandardised nature of Old French.
However, the most recent developments in the fields of lemmatisation and linguistic annotation of vernacular languages and their variety through the use of recurrent neural networks (RNN) make it possible to set up systems capable of handling the linguistic variations in historical languages\footnote{eg. Middle Dutch  \citep{kestemont_lemmatization_2016}, Medieval Latin \citep{kestemont_integrated_2017}, Medieval Occitan \citep{camps_production_2017}, Early Irish \citep{dereza2019lemmatisation},  Middle High German and a variety of other languages  \citep{manjavacas_improving_2019,schmid2019deep})} thanks to their learning and prediction capacity. Thanks to RNNs, the lemmatiser no longer needs to compare the word in the text to a list of lemmas contained in a dictionary, but is able to predict it sequence by sequence (seq2seq), character by character considering word and sentence levels. This is why we are focusing on the use of these technologies and the implementation of a lemmatisation model for Old French using the lemmatiser and POS tagger \textit{Pie} developed by E. Manjavacas \citep{enrique_manjavacas_emanjavacas/pie_2019}), which we have trained from manually annotated corpora that have been aggregated over the course of the projects.

The first experimentss on lemmatisation for Old French were launched in 2015 by Jean-Baptiste Camps and Mike Kestemont, on the texts of the \textit{Chanson d'Otinel} \citep{camps_chanson_2016}, using the newly developed Pandora tagger/lemmatiser that used convolutional and recurrent neural networks \citep{kestemont_lemmatization_2016}. It soon benefited from insertion in the broader \textsc{lakme} project (Thierry Poibeau, Daniel Stoekl, J.B. Camps et al.). Initially, the first samples of the \textit{Chanson d'Otinel} have been annotated using the TreeTagger parameters for Old French provided by Achim Stein \citep{stein_treetagger:_nodate}, then manually corrected, and used to train the first Pandora models; then, phases of data prediction, post-correction and model training were iteratively performed in order to build the corpus.
A corrected corpus of 50,000 words annotated with lemmas, morphosyntax and flexion was created for Old French and another equivalent corpus, annotated with lemmas, for Occitan (the latter in collaboration with the CORLIG project of Paris-Sorbonne) \citep{camps_production_2017}. Meanwhile, the lemmatiser Pandora was developed.

Thanks to financial support from PSL (IRIS Scripta) and above all from the DIM STCN (Ile-de-France Region), the project continued after the end of the \textsc{Lakme} project, taking the name \textsc{omél}i\textsc{e} (\textit{Outils et méthodes pour l'édition linguistique enrichie}) from 2018. It then scaled up, carrying out two tasks simultaneously: 
\begin{enumerate}
    \item the development of a lemmatisation post-correction tool, allowing fast correction and batch processing of lemmatiser output;
    \item a massive growth of the lemmatised corpus, in order to increase lemmatiser scores.
\end{enumerate}

From a computational point of view, the new annotated corpora allow, by a circular process, the training of more efficient models for lemmatisation and annotation, and thus a subsequent faster growth of the corpus.
Both the post-correction application and the lemmatiser are language-independent, and particularly suitable for non-standardised language states. The application has been continuously developed to allow collaborative work and, above all, to be used without advanced computer experience. By 2018, the post-correction application Pyrrha \citep{clerice_2021_pyrrha} was being developed and used, including for educational purposes. This was a major step forward, since this open-source service significantly improved the lemmatisation speed.

The increasing of the training corpus has made continuous progress thanks to important funding. Its cost is high and can be estimated at between 50 000 and 100 000€, not counting the development of the tools. This means that it is necessary to avoid duplicating data as much as possible and to allow interoperability of the data produced by the various lemmatisation projects in Old French or to promote data sharing or exchange. To this end, a workshop was held at the École des Chartes on 28 October 2018, dedicated to lexical lemma repositories in Old French, in order to standardise practices or at least make them interoperable. It is now necessary to develop equivalence tables between lemmas and to implement them in order to share corpora.

Methods implemented in \textsc{lakme} and then \textsc{omél}i\textsc{e} have contributed to the setting up of corpora and models for other languages, in particular for Old Occitan, but also for Classical French 
\citep{Cafieroeaax5489,camps_corpus_2020} (see also infra for the Franco-Italian).
After a complete reworking and harmonisation of the corpora, a new version of the Old French models is available since the beginning of 2021 via Pyrrha \citep{clerice_2021_pyrrha} as well as in the form of an API via Deucalion and Pie-extended \citep{pie-extended}\footnote{%
  Available at:
  \url{https://dh.chartes.psl.eu/} 
  and 
  \url{https://tal.chartes.psl.eu/deucalion/}. %
}.

\section{Corpora}
\strut


\begin{table}[htbp]
    \centering \footnotesize %
    \begin{tabular}{lp{2.5cm}p{2cm}|c|rr}
    \textbf{Dataset} & \textbf{\textit{Source}} & \textbf{Annotators} & \textbf{Morph} & \textbf{N. tokens} & \textbf{genre}\\ \hline \hline
Chrestien & Kunstmann 2009 & PK, LI & - & 252774 & romance\\ 
Code & Duval and Pastore, in progress & FC, FD, LI, NK  & partial & 160007 & law\\ 
DocLing & Glessgen 2016 & NBP, NK & full & 68317 & charters\\ 
Geste & Camps 2016 (et varia) & ACC, JBC, LI, NK & full & 195303 & epic\\ 
Lancelot & Ing (in progress) & LI & - & 286095 & romance\\ 
WauchierSConf & Pinche 2021 & AP & full & 113694 & hagiography\\ 
Varia &  &  & full & 56659 & mixed\\ \hline \hline
 \multicolumn{4}{r}{\textsc{Total}} & 1\,132\,849 & \\
    \end{tabular}
    \caption{Description of the different corpora produced by the project and used in the experiments.}
    \label{tab:corpora}
\end{table}

\subsection{Sources and history of the corpus}
\strut

The annotated texts originated, in a first time, from individual editorial (PhD) projects. The first texts to be tagged were the three manuscripts of the \textit{Chanson d'Otinel} \citep{camps_chanson_2016}, followed by other Old French Epics, now part of the \textit{Geste} database \citep{camps_geste:_2019}. The \textit{Wauchier} \citep{pinche2021edition} and \textit{Lancelot} \citep{ingIPLancelot} corpora also originate in PhD projects and are constituted of new data. 
The corpus of juridical texts results from an ongoing research project, concerning a set of vernacular translations of the \textit{Corpus juris civilis}, written between c.~1225-1275. It was also created in the context of the production of complete or partial digital editions of certain translations, in order to characterise and compare translation choices.

Apart from fully new data, preexisting datasets have also been aligned with our reference lists: it is the case of the \textit{Chrestien} corpus, originally produced under the supervision of \cite{kunstmann_chretien_2009}.

Progressively, new texts have been selected and annotated with the goal of expanding the coverage in terms of chronology, regional scriptae and genres. It resulted in the inclusion of a charters corpus, selected from the DocLing with a sampling by scripta/regional variants  \citep{glesgen_les_2016}. 
New texts keep being added progressively with this focus, with currently a corpus of lyrical poetry constituted by digitising and annotating preexisting editions
 \citep{thibaud_iv_1201-1253__comte_de_champagne_les_1925,conon_de_bethune_1160-1219_auteur_du_texte_les_1925,gace_brule_gace_1951,doss-quinby_songs_2001}
and a late allegoric verse text by Guillaume de Digulleville.

\subsection{Annotation practice and workflow}
\strut
\label{subs:annotationPractice}

The annotation workflow uses a set of tools developed as part of an initiative to establish a fully integrated environment for linguistic annotation and post-correction of historical languages.

Our annotation is characterised by its relative complexity, including the analysis of nominal and verbal flexion: pronouns are distinguished (personal, demonstrative, indefinite, interrogative, relative, adverbial, impersonal, cardinal, ordinal) and times, mode and person are specified for verbs.
Once annotated with extent models, the texts and their annotation undergo a post-correction phase inside Pyrrha \citep{clerice_2021_pyrrha}. 

Tagging post-correction can be done either linearly (from start to finish), or massively, using concordance tables (accessible from the ``Search tokens'' link). Certain categories of words are conducive to massive correction by the relative ease of identification of lemmas (NOMcom, NOMpro, Adverbe, Verbs); on the other hand, POS and morpho-syntaxic flexions always require careful analysis of syntax.
For now, inside the application {lemma+POS+flexion} are not handled as a coordinated unit, and consistency has to be verified by the human.

\subsubsection{Lemma}
\strut

The texts were annotated in lemmas, according to the entries in Tobler-Lommatzsch’s dictionary (henceforth, TL), with some adaptations.
The choice of using TL instead of, for instance, the lemmas of the \textit{Dictionnaire de Moyen Français} \citep[][henceforth DMF]{atilf_-_cnrs_dictionnaire_2015} was done bearing in mind the linguistic nature of the corpus to annotate (Old French, not Middle French) and was in line with existing tools at the time \citep[e.g.][]{stein_nouveau_nodate}. 

The reviewer verifies and manually corrects pre-annotated forms on the Pyrrha platform (and can also identify and correct at the same time the possible wrong base forms generated by the OCR/HTR of the text).
For the disambiguation of lemmas (e.g., \textit{ver1}, masc. noun, `spring' and \textit{ver2}, masc. noun, `male pig'), it is still necessary to consult the digitised TL \citep{tobler_altfranzosisches_2002} to clearly distinguish the homonyms with their definitions, because the lemmas of the TL are not intuitive and the application does not yet link to definitions. Indeed, there are homonyms even for terms that we do not think about.

For instance, in the sentence 
\begin{quotation}
\textit{El \textbf{mont} n' a home de si grant hardement}
\end{quotation}
\textit{mont} isn't ‘mountain’, but \textit{monde1} (<\textsc{mundus}), ‘world’.
Such examples are very numerous, due to the phonetic evolution of French that creates many homographs, 
e.g., \textit{errer}, \textit{errer1} < \textsc{errare}, `to make a mistake' or \textit{errer2} < \textsc{iterare}, `to go, to move'; 
\textit{mes}, \textit{mes1} < \textsc{ma(n)sum}, ‘house, garden’, \textit{mes2} < \textsc{messis}, 
‘harvest, reaping’, \textit{mes3} < \textsc{missus}, `messenger', \textit{mes4} < \textsc{missus}, `delicacies'.

The number of homograph forms is yet increased by morphological and graphetic variation, that can cause alternative spellings to collide between lemmas: e.g. \textit{mes} can be \textit{mes1}, \textit{mes2}, \textit{mes3} or \textit{mes4}, but it can also be a form of \textit{mais1} < \textsc{magis}, `more', or of \textit{mais2}, `bad', \textit{mois}, `month', \textit{manoir}, vb. `to stay', \textit{metre2}, vb. `to put, to set',
\textit{mon}, poss. `my',… 

In general, we validate or correct the lemmas proposed by the lemmatiser, but the application allows also to easily locate ``unallowed lemmas'' that do not exist in the reference list. When suitable lemmas do not exist in the TL, we create and add new lemmas. These are mostly cases of:

\begin{itemize}
    \item Proper names (first name, surname, toponym).
    \item Words not listed in the TL’s dictionary: e.g. \textit{departement} `departure'; \textit{enterinement2} `completely'.
    \item Lemmas listed in the TL, but reported at the end of the article: ex. adverbs (\textit{principaument}, \textit{mëismement}, \textit{ancïenement}, \textit{covenablement}); participles as nouns (\textit{sëu}: NOMcom, TL, s. v. \textit{savoir}), etc.
\end{itemize}

\subsubsection{Tagging principles}
\strut

POS Tagging is done following the rules given in the retained annotation scheme, \texttt{Cattex2009} \citep{guillot_manuel_2013}. Yet, some new rules had to be established and some adaptations were made, the latter especially since we are the first to use --~to our knowledge~-- the full annotation scheme, including morphological tags (\texttt{Cattex2009-full}).

\paragraph{Contraction} In the case of contractions, we have introduced dual labels; e.g., “aux” or “auquel” will be treated under “a3+le” “a3+lequel”, eg., for 
\textit{al}, \textit{au}, \textit{as} (`at the')\\

\paragraph{Agglutination/deglutination}
The division of the lexical units often differs according to the date of the text and the choice of the editor, which does not necessarily match the spacing of the manuscript; one can suppose that the script of ancient manuscripts of Old French texts is roughly more analytic than those of Middle French manuscripts. Cases include, \textit{lors que/lorsque}; \textit{toutes voies/toutesvoies}; \textit{par mi/parmi}; \textit{a fin/affin}; \textit{ja mais/jamais}, etc.  Solutions should be sought that do not unduly complicate the processing of the texts.
In case of agglutination or deglutination, we choose a solution adapted to the form of the text.
For instance, in ``Le dit jour'', if ``le dit'' is noted in 2 words, it will be treated as \verb#DETdef+VERppe#, whereas if it is noted in one word, \textit{ledit} will be tagged as a compound determinant (\texttt{DETcom}). 
However, in accordance with the lemmatisation of \cite{tobler_altfranzosisches_2002}, complex adverbial formations such as \textit{parfois}, \textit{portant}, \textit{porce}, \textit{derechief}, \textit{maintefois}, \textit{naguère}, \textit{jamais} etc. and adverbial variants with intensive prefix \textit{tres} (\texttt{ADVgen}) are always treated analytically.

\paragraph{Named entities}
The processing of named entities is left to an ulterior stage of corpus annotation. Regarding lemmatisation,  each word is tagged according to the part of the original speech, hence the fact that ``saint'', part of church names or of a toponym is treated as an adjective, e.g. ``dou dit priorei de \textit{Saint Jaike on Mont}'',

\begin{tabular}{llll}
Saint & saint & \texttt{ADJqua} &  \texttt{NOMB.=s|GENRE=m|CAS=r|DEGRE=p} \\
Jaike & Jacques & \texttt{NOMpro} &  \texttt{NOMB.=s|GENRE=m|CAS=r} \\
on & en1+le & \texttt{PRE.DETdef} &  \texttt{MORPH=empty+NOMB.=s|GENRE=m|CAS=r} \\
Mont & mont & \texttt{NOMcom} &  \texttt{NOMB.=s|GENRE=m|CAS=r} 
\end{tabular}

\paragraph{Homonyms} some very frequent homonyms deserve special attention when correcting POS and morphological tagging:
\begin{description}
\item[a] \textit{a3} (PRE), \textit{avoir1} (VERcjg), \textit{a2} (INJ);
\item[le, la, les] \textit{il} (PROper), \textit{le} (DETdef);
\item[se, s’]  \textit{soi1} (PROper), \textit{ce1} (PROind), \textit{se} (CONsub), \textit{si} (ADVgen), \textit{son4} (DETpos), \textit{ce2} (DETdem);
\item[ou, o, u]  \textit{o3} (CONcoo), \textit{où} (PROrel), \textit{en1+le} (PRE.DETdef);
\item[en]  \textit{en1} (PRE), \textit{en2} (PROadv), \textit{on} (PROind);
\item[ne] \textit{ne1} (ADVneg), \textit{ne2} (CONcoo).	
\end{description}

To this, one can add other homographs whose treatments requires a careful morpho-syntactic analysis and is tributary to the entries of the chosen lemma reference list and POS tagset \citep{tobler_altfranzosisches_2002,guillot_manuel_2013}. In particular, \textit{que} can be
\begin{description}
\item[que1 (\textsc{< quam})] CONsub (in a comparison) or ADVgen (in constructions, \textit{ne… que, ne mais que…};
\item[que2 (< \textsc{qui, quem, quam, quod})] PROrel;
\item[que3 (\textsc{< quid})] ADVint (interrogative, exclamative);
\item[que4 (< \textsc{quia})] CONsub (eg. ``il avint que'') or CONcoo (``Si sont pres de trente mile, que chevaliers que sergenz que borjois'').
\end{description}

Regarding flexion tags, there are 12 possible composite labels for all these \textit{que}. Yet its complexity is far from matching verbal forms, with about 60 possibilities (in mode, tense, person, number).

\paragraph{Irregularities, scribal or editorial mistakes, rare forms}
When a rare or apparently irregular or mistaken form is encountered, the tagging becomes more difficult.
For instance, we encounter phenomena of mismatches (in number, gender, verbal time) or case confusion, not only in versified texts (poetic license) but also in prose texts or practical documents. In this case, by examining the context, we choose the tagging that seems most consistent with the meaning:
\begin{itemize}
    \item 
\textit{Par Mahomet merveilles} (fem. pl.) \textit{ai oie} (-> fem. pl. despite the form).
\item \textit{c'est assavoir de prés, de terres que} (subj. case fem. ‘qui’) \textit{movoiet} (impft pers. 6) \textit{de moi et de mon fié, que li dis abbes avoit encloses et covert} (-> fem. pl.) en son estant dou dit priorei
\end{itemize}

Diatopic variation can also cause tagging difficulties, as regional forms create new homographs.
For instance, the Picard form of the singular feminine definite article or personal pronoun, \textit{le}, homograph to the masculine forms; the possessive \textit{se} (e.g. ``\textit{Chançon legiere a entendre Ferai,/car bien m' est mestiers /Ke chascuns le puist aprendre / Et c'on le chant volentiers}''). When we cannot decide the gender of a word, we use the value \texttt{x} especially for nouns that can be masculine or feminine, such as \textit{ost}, \textit{onor}, \textit{amour}, etc.

Pyrrha's interface \citep{clerice_2021_pyrrha} also allows to check directly the ``unallowed'' values of POS tags and, more importantly, morph (because of the way the lemmatiser is trained --~see below~--) POS tags unseen in the training set can not be predicted, while morph tags are predicted independently (i.e., genre, numb., person, case…) and then concatenated, resulting occasionally in impossible values (e.g., comparative adjective 2nd person). In all cases, linear verification is essential for lemma disambiguation and morphosyntactic analysis. Finally, to control and homogenise the tagging result, we can revise the work by concordance on the platform itself.

The current state of the system allows a fairly easy exploitation of the corpus and the tagger begins to function properly. However, it is essential to annotate the maximum number of texts now so that the tagger learns to better discern the syntax of the texts and so that we can establish its final version in an optimal form.

\section{Training setup}
\label{sec:training}
\strut

In order to get training samples 
whose structure mimics that of data observed in the real world, we ensured that our data is segmented by sentence (finishing by a \texttt{PUNfrt}) or by line (in manuscripts without punctuation). To do so, we transform the dataset using Protogenie \citep{clerice_2020_protogenie} which handles some form of normalisation: we normalise Roman numerals into Arabic numerals to reduce the complexity of the data, as well as the number of allowed numbers and we split the complex morphology into several simple categories (Case, Tense, etc.). The full corpus is then split into 3 different parts, for training, development and testing, with a 80/10/10\% ratio.

Each task is trained separately, with a fork of Pie\citep{pie}, PaPie, which provides a few additions including:
\begin{itemize}
    \item Ability to focus on various metric for improvement tracking (Accuracy, F1, Precision, Recall);
    \item New optimisers (Ranger for example) and learning rate schedulers;
    \item ``Noise introduction strategies'', such as randomised full capitalisation of sentences. 
\end{itemize}
For each task (Lemma, POS and morphology tags), we conducted a parameter research, using prior knowledge from other languages, for which we will state the value below. Each training phase was run at least 5 times to account for random local minima and each result was logged. Once all models were trained, we implemented a ranking method that ranks each model on each available and meaningful metric and chose the one whose sums of ranks is the lowest. For this, we excluded metrics such as the one applied to ``unknown targets'' based on a very small sample of lemma which would rapidly skew the overall ranking of models.

All models shared most of the same parameters: they used a single linear layer, LSTM cells for the hidden network, a character embedding using RNN, a dropout of 0.32, learning rate of 0.0049, patience for the learning rate evolution of 2, a patience for early stopping of 5. We provide the configuration on our repository. We used the Ranger optimizer that has shown less variation in training with better scores\footnote{%
    On this topic, see the discussion between \textit{TC} and Enrique Manjavacas at \url{https://web.archive.org/web/20210914113014/https://github.com/emanjavacas/pie/issues/76}.
}. 

Four parameters were used to generate diverse combinations:
\begin{itemize}
    \item The Character Embedding size, \texttt{cemb\_size}, could be of 100, 150, 200 or 300
    \item The number of layers for the Character Embeddings Encoder, \texttt{cemb\_layers}, could be 1 or 2.
    \item The size of the hidden layer, that encodes most of the context, was a value in the set $150, 200, 250, 300, 350$. We added a variation only for the lemma task at 170.
\end{itemize}

In general, while the hidden size has a low impact on the lemma task, morphological tasks were more inclined to get better results with it. The number of layers always yield better except for POS, and Nomb, indicating that the task was simpler to solve for the network at the character level, most information coming from the context. Targeting precision was efficient only for the lemmatisation task. The final best parameters are shown in table \ref{tab:configs}.

\begin{table}[t]
\centering \footnotesize %
\begin{tabular}{l|llll}
Task  & Character Layers & Embedding Size & Hidden Size  & Target    \\ \hline \hline
Lemma  & 2                & \textbf{300}   & \textit{150} & Precision \\ \hline
POS   & 2                & 200            & \textbf{350} & Accuracy  \\ \hline
CAS   & 2                & 150            & \textit{150} & Accuracy  \\
DEGRE & 2                & 200            & 250          & Accuracy  \\
GENRE & 2                & 200            & 250          & Accuracy  \\
MODE  & 2                & \textit{150}   & 200          & Accuracy  \\
NOMB  & 1                & 200            & 250          & Accuracy  \\
PERS  & 2                & \textit{150}   & \textbf{350} & Accuracy  \\
TEMPS & 2                & 200            & \textbf{350} & Accuracy         
\end{tabular}
\caption{Best parameters found after the sweep. Bold are the highest value, italic the lowest.}
\label{tab:configs}
\end{table}

\label{sec:training-lemmatisation}

\section{Results}
\strut

\subsection{Scores}

\begin{table}[h!]
    \centering \footnotesize %
\begin{tabular}{l||rrr|rrr|rrr|rrr}
      & \multicolumn{3}{c|}{All} & \multicolumn{3}{c|}{Known tokens} & \multicolumn{3}{c|}{Unknown tokens} & \multicolumn{3}{c}{Ambiguous tokens} \\
 task &   acc &   pre &   rec &          acc &   pre &   rec &            acc &   pre &   rec &              acc &   pre &   rec \\ \hline \hline
lemma & 97.66 & 76.08 & 75.86 &        98.38 & 91.72 & 91.71 &          69.01 & 48.86 & 48.72 &            97.79 & 78.24 & 78.11\\ \hline
 POS & 97.55 & 83.00 & 79.90 &        97.88 & 84.81 & 82.11 &          84.56 & 54.19 & 55.92 &            97.37 & 83.70 & 81.20 \\ \hline
  CAS & 95.25 & 91.75 & 92.26 &        95.74 & 92.18 & 92.74 &          84.63 & 62.60 & 62.55 &            94.01 & 91.77 & 92.33 \\
DEGRE & 98.47 & 86.45 & 82.77 &        98.74 & 87.75 & 84.98 &          92.55 & 63.81 & 44.48 &            95.46 & 86.91 & 85.80 \\
GENRE & 96.21 & 93.22 & 89.03 &        96.76 & 93.67 & 89.66 &          84.47 & 88.01 & 69.87 &            94.59 & 92.11 & 87.90 \\
 MODE & 99.05 & 91.59 & 89.00 &        99.36 & 93.56 & 92.06 &          92.39 & 79.46 & 74.89 &            97.61 & 88.89 & 92.14 \\
 NOMB & 97.05 & 96.50 & 96.20 &        97.47 & 96.92 & 96.65 &          88.20 & 87.85 & 82.20 &            96.06 & 95.67 & 95.36 \\
 PERS & 98.79 & 91.25 & 85.28 &        98.95 & 91.73 & 85.73 &          95.34 & 90.17 & 90.12 &            97.72 & 91.31 & 85.12 \\
 TEMPS & 99.13 & 96.51 & 96.20 &        99.38 & 97.21 & 97.58 &          93.94 & 88.92 & 85.39 &            97.80 & 95.02 & 96.78
\end{tabular}
    \caption{Lemmatisation and tagging accuracies on the test set for the best model for each configuration. ``Unknown tokens'' are tokens never seen during training, while ``ambiguous tokens'' are forms that can correspond to different lemmas. Results for ``Unknown targets'' (lemmas never seen in training, but that the neural network can still accurately predict thanks to its character level modelling) are not given because the support (258 tokens) was too low for it to be significant.}
    \label{tab:pie_accs}
    \end{table}

Scores are shown in Table~\ref{tab:pie_accs}. They reach a global level of accuracy that is similar regarding lemma and POS (c.~97.5\%), and more heterogeneous  for morphological tags, from 95.25\% for case to 99.05\% for verbal mode.
The accuracies are relatively robust to ambiguous tokens, at least for lemma and POS. 

Thanks to its character level modelling, the model also achieve a \textbf{69\% accuracy for lemma prediction on tokens never seen during training}.

It is to be noted that, behind the composite figure of 97.66\% accuracy for lemmas, variation are to be found between grammatical categories,
with 100\% accuracy being reached for unambiguous and easy to tag punctuation signs or possessive adjectives (\textit{mien}, \textit{nostre}, etc.)
and relatively low scores for misspelled words  (\texttt{OUT}), interjections, categories with a grammatical ambiguity or proper names (see in appendix~\ref{appendix:tab:lemmaPerPos} for detailed scores). 

Some lower scores are affected by grammatical ambiguity and arbitrary lemmatisation choices. 
Apart from the different forms of \textit{que}, that we will discuss in next subsection, the adjectival forms of the verbs and the qualifying adjectives have score that are driven down by this. 
For instance, for \texttt{VERppa}, the choice of tags for the forms in -\textit{ant} (\textit{vaillant}, \textit{chantant}, \textit{vivant},…), that in Modern French, can be depending on the case, either adjectives or present participles, but are
not so strictly classified in Old French, necessitate a more theoretical and arbitrary decision than some other categories. They
are here tagged (according to \texttt{Cattex2009} principles, \cite{prevost_principes_2013}) always as verbal forms (\texttt{VERppa}) with a human annotated verbal infinitive as lemma (\textit{vaillant}$\rightarrow$\textit{valoir}, \textit{chantant}$\rightarrow$\textit{chanter}, etc.), yet the tagger is on occasion inclined to treat them as adjectives (indeed close to their actual syntactic function).

\subsection{Errors and Most frequent confusions}

To go beyond the constructed numeric values of scores, it is possible to inspect the confusion matrix indicating the most frequent types of wrong predictions and a classical tool in machine learning (Tables~\ref{tab:confMatlemmas} and \ref{tab:confMatPOS}).

\begin{table}[htbp]
\centering \footnotesize %
\begin{tabular}{lrlr}
 \textbf{GT} & \textbf{Errors} & \textbf{Preds} & \textbf{Freq} \\ \hline \hline
 que2 & 147 & que4 & 137 \\ 
 que4 & 50 & que2 & 41 \\ 
 que1 & 49 & que4 & 46 \\ 
 il & 42 & le & 36 \\ 
 le & 26 & il & 24 \\ 
 a3 & 21 & avoir & 19 \\ 
 ne1 & 20 & ne2 & 20 \\ 
 avoir & 19 & a3 & 13 \\ 
 que3 & 16 & que4 & 10 \\ 
 ne2 & 15 & ne1 & 15 \\ 
 si & 14 & se & 9 \\ 
 se & 13 & soi1 & 6 \\ 
 on & 13 & en3 & 5 \\ 
 en2 & 12 & en1 & 6 \\ 
 ce2 & 11 & ce1 & 6 \\ 
    \end{tabular}
        \caption{Confusion matrix for lemmas (only values $> 10$ were kept).}
    \label{tab:confMatlemmas}
\end{table}

\begin{table}[!htb]
\centering \footnotesize %
\begin{tabular}[t]{lrlr}
 \textbf{GT} & \textbf{Errors} & \textbf{Preds} & \textbf{Freq}  \\ \hline \hline
 NOMcom & 203 & ADJqua & 53 \\ 
 & & VERppe & 38 \\ 
 & & VERcjg & 31 \\ 
 & & ADVgen & 25 \\ 
 & & NOMpro & 19 \\ 
 & & VERinf & 15 \\ 
 ADVgen & 173 & NOMcom & 35 \\ 
 & & CONsub & 26 \\ 
 & & PRE & 21 \\ 
 & & PROind & 16 \\ 
 & & ADJqua & 15 \\ 
 & & PROper & 13 \\ 
 VERcjg & 155 & VERppe & 59 \\ 
 & & NOMcom & 44 \\ 
 & & PRE & 21 \\ 
 ADJqua & 119 & NOMcom & 66 \\ 
 & & VERppe & 23 \\ 
 CONsub & 119 & PROrel & 83 \\ 
 & & ADVgen & 24 \\ 
 PROrel & 98 & CONsub & 86 \\ 
 PRE & 78 & ADVgen & 37 \\ 
 & & VERcjg & 15 \\
 & & PROadv & 10 \\
  VERppe & 78 & VERcjg & 38 \\ 
 & & NOMcom & 16 \\ 
 & & ADJqua & 14 \\
 PROimp & 78 & PROper & 78 
 \end{tabular} %
     \begin{tabular}[t]{lrlr}
 \textbf{GT} & \textbf{Errors} & \textbf{Preds} & \textbf{Freq}  \\ \hline \hline
 OUT & 70 & VERcjg & 13 \\ 
 PROper & 69 & DETdef & 22 \\ 
 & & PROimp & 17 \\ 
 & & DETpos & 15 \\ 
 PROind & 64 & DETind & 22 \\ 
 & & ADVgen & 18 \\ 
 NOMpro & 48 & NOMcom & 20 \\ 
 CONcoo & 45 & ADVneg & 26 \\ 
 DETdef & 36 & PROper & 31 \\ 
 DETcar & 28 & DETndf & 12 \\ 
 & & ADJcar & 10 \\ 
 ADVneg & 25 & CONcoo & 14 \\ 
 ADJpos & 25 & DETpos & 18 \\ 
 DETind & 23 & PROind & 8 \\ 
 VERinf & 22 & NOMcom & 15 \\ 
 DETpos & 22 & PROper & 5 \\ 
 PROcar & 21 & DETcar & 6 \\ 
 ADJcar & 20 & DETcar & 10 \\
ADJind&19&DETind&13\\ 
VERppa&17&ADJqua&6\\ 
ADVint&17&CONsub&9\\ 
PROadv&15&PRE&6\\ 
PROint&14&PROrel&9\\ 
PROdem&13&DETdem&9\\ 
PROord&12&ADJord&8\\ 
ADVsub&11&CONsub&4\\ 
DETdem&10&PROdem&6              
\end{tabular}
    \caption{Confusion matrix for parts-of-speech (only values $\geq 10$ were kept).}
    \label{tab:confMatPOS}
\end{table}

Concerning lemmas, two major conclusions can be drawn:
\begin{enumerate}
    \item the most frequent errors are massively related to homographs, and particularly homographs of function words (\textit{que1-4}, \textit{ne1-ne2}, \textit{le} as pronoun or definite determiner, etc.),  
    that also cause trouble to human annotators (see above, subsection~\ref{subs:annotationPractice}). 
    Particularly, confusions between \textit{que1}, \textit{que2}, \textit{que3} and \textit{que4} account for 15.25\% of all errors.
    \item the most frequent errors are on function words, which is easily explainable given that they form a large majority of the total number of words for a small number of lemmas,
    \item meanwhile, \textbf{around 40\% of errors are distributed between 679 different lemma with a single error} (e.g., \textit{ avoutire1}, \textit{hustin}, \textit{Mordred}, …), mostly proper names, content words and rare forms (which are interesting to lexicographers),
    \item and a majority of errors themselves (e.g., \textit{il} instead of \textit{avoir}, \textit{venir} instead of \textit{aler}, etc.) are in single occurrence (53\%), which complicates the task of batch correction.
\end{enumerate}

Concerning part-of-speech tags (see detailed scores per category in appendix, Table~ \ref{appendix:tab:posScores}), 
a few categories achieve a perfect score due to their mostly unambiguous nature and very low diversity, such as punctuation signs (\texttt{PON...}),
while, at the other end of the spectrum 
the lowest scores are reached for rare and ambiguous categories like the homographic impersonal pronoun \textit{il} (\texttt{PROimp}, vs \textit{il} \texttt{PROper}), the homographic 
\textit{que3}, \textit{come1} and other \texttt{ADVint}, or the category \texttt{OUT}, that was left in the test set, but whose presence is arguable, because this category concerns words that are taken out of grammatical analysis (mostly scribal mistakes, for instance, such as words repeated a second time or left unfinished by the scribe). 
It is to be noted that these low F1 scores\footnote{%
   The F1 score is the harmonic mean of precision and recall, and as such a global measure of accuracy. 
} are mostly driven down by a very low recall. It can be interpreted as a large tendency of the tagger to give to occurrences of this rarer categories for a given form or lemma the tags from one more often encountered for them in the training material, as can also be seen from the confusion matrix (Table~\ref{tab:confMatPOS}).

Content-words categories (such as \texttt{NOM...}, \texttt{ADJ...} or \texttt{VER...}) achieve relatively high scores, yet their high frequency makes them featuring prominently in the confusion matrix, with some of the most frequent confusions being between common nouns and adjectives, or between the different verbal subcategories.

\subsection{Qualitative inspection of lemmatisation results and comparisons with other models}

For a more direct inspection by the human, 
we evaluate the results of the lemmatiser on a test set, which contains 2\,849 
random sentences drawn from the corpora. The evaluation is made on the comparison between the prediction of the model and the manually corrected gold standard. We then compute a sentence-level word-based accuracy (number of words with at least one error divided by total number of words). 
The distribution of sentences scores is presented %
in Table~\ref{tab:sentencesScores}.

\begin{table}[!ht]
    \centering
    \begin{tabular}{c|c}
        Sentence score & Number of sentences  \\
        \hline
        1 &  1\,871\\
        0.9 -- 1 & 842\\
        0.8 -- 0.9 & 114\\
        < 0.8 & 22\\
    \end{tabular}
    \caption{Number of sentences for each score}
    \label{tab:sentencesScores}
\end{table} 

The results show that most sentences are correct (around 95\%).
The sentence with the worst score (0.32) is actually not written in Old French but in Latin. Two other sentences have a very weak score, of 0.4 (Table~\ref{tab:worstSents}; from now on, errors are shown in italics in the tables).

\begin{table}[!h]
    \centering \footnotesize %
    \begin{tabular}{c||ccccc}
    tokens &  Qui & montagu & auoit & a & iustisier\\
    correct   & qui & Montagu & \textit{aler} & a3 & justicier2\\
    predicted & qui & \textit{montaignor} & avoir & a3 & \textit{vistoier}\\ 
    TT-TL & \textit{cuidier1} & \textit{montagu} & avoir & \textit{a3|a} & \textit{justicier2|justicier}\\
    TT-MF & qui & \textit{montagu} & avoir & a & \textit{iustisier}\\
    \hline\hline
    tokens & A & Gironuille & uont & ludie & veir
      \\
    correct & a3 & Gironville & aler & Ludie & vëoir\\
    predicted & a3 & \textit{Gironle} & \textit{avoir} & Ludie & \textit{vair1}\\
    TT-TL & \textit{avoir} & \textit{Gironuille} & aler & \textit{ludie} & \textit{vair1|voir}\\
    TT-DMF & a & \textit{Gironuille} & aller & \textit{ludie} & \textit{voir}\\
    \end{tabular}
   \caption{Sentences with worst scores in the test set; row gives the original tokens, the human annotated ground truth (correct), our model prediction (predicted), and, for comparison, the results of two sets of parameters for the TreeTagger lemmatiser.}
   \label{tab:worstSents}
\end{table}

Two parameters explain the weakness of the score: first, the sentences are short, so one error strongly affects the score; secondly, these sentences present peculiar spellings, contrary to the majority of the tokens in the corpus, either because of diatopic variation or editorial choices (e.g., distinction \textit{i}/\textit{j} and \textit{u}/\textit{v}). In this particular example, we note several difficulties. One concerns the proper nouns, with \textit{montagu} interpreted as \textit{montaignor} and \textit{Gironuille} as \textit{Gironle}. An other one is the problem of multiple spellings, as \textit{veir} as an occurrence of \textit{vëoir} is difficult to identify. We also note that the human corrected value can sometimes be the wrong one, as \textit{auoit} was wrongly identified by the human annotator as an occurrence of \textit{aler}. The lemmatiser, here, is right. This kind of error can happen elsewhere (Table~\ref{tab:lemmatiserBetter}).

\begin{table}[!h]
    \centering \footnotesize 
    \begin{tabular}{c||cccccccc}
        tokens &  Et & bien &sachiez & qu' & en& l' & eglisse\\
        correct & et & bien1 & \textit{sachier2} & que4 & en1 & le & eglise\\
        predicted & et & bien1 & savoir & que4 & en1 & le & eglise\\
   \end{tabular}
    \caption{lemmatiser can perform better than the human annotator.}
   \label{tab:lemmatiserBetter}
\end{table}

This kind of error tends to show that the score that we present is at times lower than it should be.
Moreover, other frequent errors which bring down the sentence scores happen because of annotators tag choice regarding capitalisation, numerals, or are due to the tags of the interjections, all that may be at times inconsistently tagged by human annotators (Table~\ref{tab:tagChoice}).

\begin{table}[!h]
    \centering \footnotesize %
    \begin{tabular}{c||c|c||c||c|c}
        tokens & sarrasins & dex & .l.m. & O & Ha\\
        correct & sarrasin & dieu & 50000 & o2 & \textit{a2}\\
        predicted & Sarrasin& Dieu& \textit{50} & \textit{ho!} & ha!
    \end{tabular}
    \caption{Problems with the choice of the tag}
   \label{tab:tagChoice}
\end{table}

All of these are marginal errors, that could be resolved with some additional work on the  harmonisation of the corpora. Some errors are more important. They are caused by homography and often concern personal pronouns and possessive, as we can see in Table~\ref{tab:pbPPP}.

\begin{table}[!h]
    \centering
    \footnotesize
    \begin{tabular}{c||cccp{0.65cm}p{0.75cm}cccccc}
        tokens & Et & veez & la & la &. &&&&&& \\
        correct & et & vëoir & il & là & .&&&&&&\\
        predicted & et & vëoir & \textit{là} & là &. &&&&&&\\
        TT-TL& et & \textit{vëer|vëoir} & \textit{le} & \textit{là|il} & . &&&&\\
        TT-DMF & et & vëoir & \textit{là} & là &. &&&&&&\\
        \hline\hline
        tokens & et & nos & partimes &&&&&&&&\\
        correct & et & nos1 & partir&&&&&&&& \\
        predicted  & et & \textit{nostre} & partir &&&&&&&&\\
        TT-TL & et & \textit{nos|nos1|noz1} & partir\\
        TT-DMF & et & nos & partir\\
                \hline\hline
        tokens & li & reis & garsie & est& mis & germeins & cusins & mis & uncle & fu & fernagu\\
        correct & le & roi & Garsie & estre1 & mon1 & germain & cosin & mon1 & oncle & estre1 & Fernagu\\
        predicted  & le & roi & Garsie & estre1 & \textit{metre2} & germain & cosin & mon1 & oncle & estre1 & Fernagu\\
        TT-TL & le & roi & \textit{garsie} & \textit{ester}|\newline \textit{estre1} & \textit{manoir}|\newline \textit{metre}|\newline \textit{mettre} & \textit{<nolem>} & cosin & mon1 & oncle & \textit{estre1|estre} & \textit{fernagu}\\
        TT-DMF & le & roi & Garsie & estre1 & \textit{metre2} & \textit{germeins} & cosin & mon1 & oncle & estre1 & Fernagu
    \end{tabular}
    \caption{Problems with personal pronouns and possessives}
    \label{tab:pbPPP}
\end{table}

The errors are understandable: the form \textit{la} can indeed
refer to the adverb \textit{là}, in the first sentence, as the form \textit{nos}, in the second one, can be an occurrence of the personal pronoun or of a possessive. Interesting is the form \textit{mis}, which can be an occurrence of \textit{metre2}, which is once predicted wrong, and once right, in the same sentence.

Other errors which are caused by the existence of similar spellings can be more important (Table~\ref{tab:dessirier}). Here, the verb \textit{descirier}, ``to tear'', is predicted as \textit{desirrier2}, a substantive meaning ``whish''. This error is important because it changes the semantics of the whole sentence.

\begin{table}[!h]
    \centering
    \begin{tabular}{c||cccccc}
        tokens & Et & tant & mantel & desrompre & et & dessirier \\
        correct & et & tant & mantel & desrompre & et & descirier\\
        predicted & et & tant & mantel & desrompre & et & \textit{desirrier2} \\
        TT-TL & \textit{avoir} & tant & mantel & desrompre & et & descirier\\
        TT-DMF & et & tant & manteau & \_ & et & déchirer
    \end{tabular}
    \caption{Wrong prediction due to homography}
    \label{tab:dessirier}
\end{table}

We can compare the results of the lemmatiser with the ones of an other one, TreeTagger, for which at least two sets of parameters for Old French exist,
using the lemmas from TL 
\citep{stein_treetagger:_nodate}, henceforth \textbf{TT-TL} 
or from the DMF \citep{BFM:_nodate}, henceforth \textbf{TT-DMF} 
\footnote{%
    They are available at:
    \url{https://sites.google.com/site/achimstein/research/resources} and 
    \url{https://www.cis.lmu.de/~schmid/tools/TreeTagger/}.
}.
The results for the two sentences with low scores tagged with TreeTagger are shown in Table~\ref{tab:worstSents}.

The lemmatiser has the same difficulties which were mentioned above: proper nouns and particular spellings (\textit{veir}). If its proposition for the form \textit{iustisier} is better than the one from our model, it makes mistakes for easier occurrences (\textit{Qui}, relative pronoun, is identified as an occurrence of \textit{cuidier1}, ``to believe''; \textit{A}, preposition, is identified as an occurrence of the verb \textit{avoir}). It also encounters difficulties with personal pronouns and possessive and produces the same errors as we saw in table \ref{tab:pbPPP}.

 The TT-TL and TT-DMF parameters produce one error our model didn't produce: it cannot identify the form \textit{germeins}. The TT-TL parameters, on the test set, provides for 1,561 occurrences a \textit{<nolem>} tag, and the TT-DMF ones, a ``\_'' tag, as in table \ref{tab:dessirier}, for 2,878 occurrences. The TT-TL parameters also propose alternative choices where our model proposes just one.
    
On occasions, our model is seen better handling spellings marked from a diachronic or diatopic perspective, or more generally, less familiar spellings. For instance, it handles better the Anglo-Norman forms seen in \ref{tab:archaicForms}, where TT-TL makes six mistakes, TT-DMF makes five and our model none (but it is a sample drawn from an in-domain text).
  
\begin{table}[!h]
\centering  \footnotesize %
 \begin{tabular}{c||cccccccccccc}
tokens  & laisum & clarel & cest & saracin & aler& , & kar &bin & vez & nel &pouum& mener \\
correct & laissier & Clarel & cest & Sarrasin & aler & , & car & bien1 & vëoir & ne1+il & pöoir & mener\\
predicted &  laissier & Clarel & cest & Sarrasin & aler & , & car & bien1 & vëoir & ne1+il & pöoir & mener\\
TT-TL & laissier & \textit{clarel} & cest & saracin & \textit{aler|foraler} &,& car & \textit{<nolem>} & \textit{aler|vëoir} & \textit{il|ne1} & pouvoir & mener\\
TT-DMF & laisser & \textit{clarel} & cest & saracin & aller & , & car & \textit{bin} & \textit{fois} & ne.il & \textit{pouum} & mener
 \end{tabular}
 \caption{Comparisons of the results on one sentence with Anglo-Norman forms}
  \label{tab:archaicForms}
 \end{table}

It can happen that the results of TreeTagger are better than ours, for instance, in the sentence with the difficult form \textit{dessirier} (Table~\ref{tab:dessirier}).
However, TT-TL makes a mistake for the identification of the easy form \textit{Et}, which cannot be identified, perhaps due to capitalisation. Moreover, it on turns produces hesitations or mistakes on similar forms (Table~\ref{tab:dessirant}).

\begin{table}[!h]
    \centering \footnotesize %
    \begin{tabular}{c||cccccccc}
          tokens &As & paiens & sont & venu & ,& de & ferir & desirant  \\
      correct & a3+le &  paien & estre1 & venir &,  & de  & ferir & desirrer\\
      predicted & a3+le &  paien & estre1 & venir&, & de  & ferir & desirrer\\
      TT-TL & \textit{a+le|le} & paien & \textit{estre1|estre} & venir &,& de & ferir & \textit{descirier|desirer}\\
      TT-DMF   & a.le &païen & être & venir &, &de &férir& \textit{déchirer}\\
    \end{tabular}
    \caption{Prediction of TreeTagger on the form \textit{desirant}}
        \label{tab:dessirant}
\end{table}

\section{Discussion and further research}
\strut

Current models display very satisfying results, in particular when accounting for the specific difficulties of an under-resourced variation-rich non-standardised historical language, in which variation also increases the number of homograph forms and the size of the ``vocabulary'' of forms. The use of a neural tagger and the creation of a significant gold corpus has allowed to obtain very usable results, especially when tested on in-domain material.

In some cases, it might be possible to achieve further gains in accuracy by crossing predictions for lemma and for part-of-speech (which are for now predicted independently by the neural network). This could be the case, for instance, for the various occurrences of \textit{que<n>} (Table~\ref{tab:betterQue}).

\begin{table}[htbp]
    \centering \footnotesize %
\begin{tabular}{lrrrrrrr}
 GT   &   Freq &   Acc. POS &   Acc. Lemma &   Combined Accs &   Pred$=$\textit{que4} &   Pred=\textit{que2} \\ \hline
\hline
 que1&97&\textbf{82.47}&48.45&39.18&48.45&3.09 \\
 que2&587&\textbf{84.67}&75.98&72.57&22.49&75.98 \\
 que3&37&\textbf{67.57}&59.46&56.76&24.32&16.22 \\
 que4&1474&92.13&\textbf{96.40}&90.84&96.40&2.92
\end{tabular}
    \caption{Lemma and POS accuracy of the prediction for the different kinds of \textit{que}; better performance of the POS indicates that in many cases lemmatisation accuracy could be increased through post-treatment after the lemma prediction phase.}
    \label{tab:betterQue}
\end{table}

Nonetheless, the generality of the models will need to be evaluated in a more systematic fashion, regarding the variety of regional scriptae and written genres of Old French. 
For this, we will need to build an out-of-domain corpus
to get a more general evaluation of the model, and to test its performances on  specific scriptae or genre (What are the performances on Picard or Lorrain texts? On theatre? etc.).

Another interesting lead is to explore transfer learning approaches on neighbouring languages, such as Occitan for which some early experiments have been conducted \citep{camps_production_2017}, 
on later varieties such as Middle and Pre-Classical French, 
and on Franco-Romance hybrids, in particular Franco-Italian, for which the creation of an annotated corpus is ongoing \citep{gambino_rialfri_nodate,ceresato_entree_tobepubl}.

Diachrony should also be explored. For the moment, 
For the moment the corpus includes only one text from the 14th century, an extract from the \textit{Pèlerinage de l'âme} by Guillaume de Digulleville (1355-1358). Given the costs of building lemmatised corpora, one must be careful not to create duplicates. For this reason, an extension towards Middle French would need to be done in order to maintain interoperability of the corpora with the \textit{Dictionnaire de Moyen Français} and the tools developed around it.
Integrating this period would fill the current break between
the models we make available for Old French and for Early Modern French \citep{camps_corpus_2020} in order to allow for studies in long diachrony \citep{atilf_-_cnrs_dictionnaire_2015}. 

More generally, further work needs to investigate the interoperability of annotation schemes. One possibility could be to offer a version of the corpora converted to the use of Universal POS tags (UD POS), that are of broad use today for many Contemporary languages.
In any case, there is a need for the elaboration of a common canonical reference tag-set and lemma list, to which and from which designing conversions for all the formats that are currently being used by the projects working on Old French lemmatisation.

\section*{Authors contributions}
\strut

\textbf{JBC} started and over-viewed the projects, trained some of the models, took part alone or with others in the annotation or correction of the corpora Geste, worked on the workflows and reference lists; he coordinated the team, with FD.
\textbf{TC} worked mostly on the engineering and NLP research around this corpus, built the interfaces that are used (Pyrrha, Deucalion), organized the training and optimization of the models, expanded tools to ensure the applicability outside of tests (PaPie) and engineered the various tools that ensure reproducibility and quality control (Protogenie, PyrrhaCI). He provided some insights when necessary regarding some annotation choices and technological limitations.
\textbf{FD} worked on the post-correction of juridical texts and coordinated the team with JBC.
\textbf{LI} joined the project during its early phase and worked on the correction of the texts in the \textsc{lakme} and \textsc{omél}i\textsc{e} projects; she also  provided data from her PhD (in progress).
\textbf{NK} worked on the post-correction (lemma, POS and morphosyntactic labelling) of texts that were used to improve automated tagging performance.
\textbf{AP} joined JBC in annotating data in parallel of the \textsc{lakme} project with data from her PhD Thesis, provided most of the use-cases and early tests of Pyrrha by correcting her corpus (lemma and morpho-syntax). 

The authors have no competing interests to declare.

\section*{Materials and data availability}
\strut

The most up-to-date version of the models can be easily obtained and used thanks to the \texttt{pie-extended} Python package, available on Pypi (\url{https://pypi.org/project/pie-extended/}), with the command \texttt{pie-extended download fr}, and can be queried at \url{https://tal.chartes.psl.eu/deucalion/}.

\section*{Acknowledgements}
\strut

This work benefited from several projects funded by various institutions: \textsc{lakme} -- \textit{Linguistically Annotated Corpora Using Machine Learning Techniques} (2016-2018, Université PSL); \textsc{omél}i\textsc{e} -- \textit{Outils et méthodes pour l'édition linguistique enrichie} (2018-…, Université PSL/IRIS Scripta and Région Île-de-France, DIM \textit{Sciences du texte et connaissances nouvelles}), as well as, for the legal corpus, of \textit{Biblissima}.
We thank the DIM \textit{Science du texte et connaissances nouvelles} for funding the acquisition of a GPU server, as well as the École nationale des chartes for providing infrastructure and support for the server.

We thank Mike Kestemont for the collaboration regarding neural lemmatisers, as well as Enrique Manjavacas for his precious advice regarding lemmatisation and Pie configuration. 
We also thank Thierry Poibeau (LATTICE) and Daniel Stoekl (EPHE), with which the initial \textsc{lakme} project was launched. 
Our gratitude for the collaboration also goes to the team working on Franco-Italian texts, led by Francesca Gambino. 
It is difficult to acknowledge the help of all colleagues that helped along the years, but we non exhaustively thank 
Floriana Ceresato, Simon Gabay, Sophie Prevost as well as all the participants in the workshop \og{}Référentiels de lemmes du français médiéval\fg{}, in particular, Alexei Lavrentiev, Martin Glessgen, Simon Gaunt, Maud Becker and Gilles Souvay, without forgetting the students of the École des chartes.
\bibliographystyle{plainnat}
\bibliography{corpusAndModels}

\appendix\footnotesize

\clearpage

\section{Detailed scores per part-of-speech}
    \label{appendix:tab:posScores}
\strut

\begin{table}[!hbp]
    \centering \footnotesize
\begin{tabular}{l|rrrrr}
 \textbf{target}&\textbf{precision}&\textbf{recall}&\textbf{f1-score}&\textbf{support} \\ \hline \hline
 PONfbl&1.00&1.00&1.00&4746\\ 
 PONfrt&1.00&1.00&1.00&2764\\ 
 PONpdr&1.00&1.00&1.00&450\\ 
 PONpga&1.00&1.00&1.00&766\\ 
 PROper.PROper&1.00&1.00&1.00&13\\ 
 ADVneg.PROper&1.00&0.98&0.99&54\\ 
 VERcjg&0.99&0.98&0.99&9745\\ 
 CONcoo&0.99&0.99&0.99&3982\\ 
 DETdef&0.99&0.99&0.99&3662\\ 
 PRE&0.99&0.99&0.99&5618\\ 
 PRE.DETdef&0.99&0.99&0.99&979\\ 
 PROdem&0.99&0.99&0.99&913\\ 
 ADVneg&0.98&0.98&0.98&1543\\ 
 NOMcom&0.98&0.98&0.98&10007\\ 
 PROadv&0.98&0.98&0.98&811\\ 
 PROper&0.98&0.99&0.98&6168\\ 
 VERinf&0.98&0.99&0.98&1537\\ 
 DETpos&0.97&0.98&0.98&1361\\ 
 NOMpro&0.98&0.97&0.97&1715\\ 
 ADVgen&0.97&0.97&0.97&4999\\ 
 INJ&0.97&0.97&0.97&33\\ 
 DETdem&0.96&0.98&0.97&413\\ 
 DETndf&0.95&0.98&0.96&403\\ 
 CONsub&0.95&0.95&0.95&2578\\ 
 VERppe&0.94&0.96&0.95&2213\\ 
 ADVgen.PROper&0.91&1.00&0.95&10\\ 
 ADJqua&0.94&0.93&0.94&1800\\ 
 PROrel&0.93&0.94&0.94&1630\\ 
 DETind&0.92&0.96&0.94&591\\ 
 PROind&0.93&0.91&0.92&681\\ 
 PRE.PROdem&0.88&0.93&0.90&15\\ 
 ADJind&0.89&0.89&0.89&171\\ 
 VERppa&0.89&0.87&0.88&134\\ 
 DETcar&0.87&0.84&0.85&171\\ 
 PROint&0.91&0.78&0.84&65\\ 
 PROpos&0.84&0.84&0.84&43\\ 
 ADJord&0.76&0.92&0.83&37\\ 
 ADVsub&0.85&0.79&0.82&52\\ 
 PROcar&0.82&0.79&0.81&101\\  
 DETrel&0.83&0.68&0.75&22\\ 
 ADJcar&0.77&0.73&0.75&75\\ 
 DETint&0.71&0.77&0.74&13\\ 
 ADJpos&0.84&0.62&0.71&66\\ 
 PROord&1.00&0.52&0.68&25\\ 
 ADVint&0.78&0.55&0.65&38\\ 
 OUT&0.91&0.49&0.64&137\\ 
 PROimp&0.80&0.46&0.59&145\\ 
\end{tabular}
\caption{%
Detailed scores for each part-of-speech in the test set, giving precision, recall, F1 and support. 
POS with total frequency $<$ 10 were removed (PRE.DETrel, PRE.PROper, DETcom, DETord, RED).
}
\end{table}

\clearpage

\section{Lemma scores for each part-of-speech category}

\label{appendix:tab:lemmaPerPos}
\strut

\begin{table}[!hbp]
    \centering \footnotesize
\begin{tabular}{l|rrrrr}
 \textbf{POS}           &   \textbf{Lemmas Accuracy} &   \textbf{Freq} &   \textbf{Lemmas SDI} \\
\hline \hline
 PONfrt&100.00&2764&0.34 \\
 PONpga&100.00&766&0.00 \\
 PROimp&100.00&145&0.00 \\
 ADJpos&100.00&66&1.87 \\
 ADVneg.PROper&100.00&54&0.00 \\
 DETrel&100.00&22&0.30 \\
 PRE.PROdem&100.00&15&0.00 \\
 DETint&100.00&13&0.00 \\
 PROper.PROper&100.00&13&0.00 \\
 ADVgen.PROper&100.00&10&0.00 \\
 PONfbl&99.96&4746&0.55 \\
 DETind&99.66&591&1.87 \\
 ADJind&99.42&171&1.28 \\
 DETdef&99.34&3662&0.01 \\
 PONpdr&99.33&450&0.00 \\
 CONcoo&99.27&3982&0.99 \\
 DETndf&99.26&403&0.00 \\
 PRE&99.07&5618&2.19 \\
 PROper&98.96&6168&1.30 \\
 DETpos&98.90&1361&1.48 \\
 PRE.DETdef&98.88&979&1.00 \\
 PROdem&98.58&913&0.81 \\
 ADVneg&98.38&1543&0.92 \\
 PROadv&98.27&811&0.65 \\
 VERcjg&97.76&9745&3.96 \\
 ADVgen&97.76&4999&3.83 \\
 VERinf&97.46&1537&4.99 \\
 ADJord&97.30&37&1.87 \\
 DETdem&97.09&413&1.16 \\
 PROind&97.06&681&2.35 \\
 NOMcom&96.80&10007&6.10 \\
 ADVsub&96.15&52&0.70 \\
 PROcar&96.04&101&2.82 \\
 ADJcar&96.00&75&2.03 \\
 CONsub&95.62&2578&1.26 \\
 VERppe&95.62&2213&5.42 \\
 ADJqua&95.56&1800&4.31 \\
 PROpos&93.02&43&1.48 \\
 PROord&92.00&25&1.91 \\
 NOMpro&91.49&1715&5.17 \\
 PROint&90.77&65&1.52 \\
 PROrel&90.18&1630&1.33 \\
 DETcar&90.06&171&3.11 \\
 VERppa&88.06&134&3.90 \\
 INJ&84.85&33&1.68 \\
 ADVint&78.95&38&1.15 \\
 OUT&62.04&137&1.40
\end{tabular}
\caption{%
Detailed scores of the lemmas for each part-of-speech, giving lemma accuracy, total frequency of the tag, and the Shannon Diversity Index (0 means no diversity). 
POS with total frequency $<10$ were removed (PRE.DETrel, PRE.PROper, DETcom, DETord, RED).
}
\end{table}

\end{document}